\documentclass[12pt]{article}
\usepackage{graphicx,psfrag,epsf}
\usepackage{enumerate}
\usepackage{natbib}
\usepackage{url} 
\usepackage{hyperref}
\usepackage{amsmath,amssymb,amsfonts,bm,bbm,amsthm} 
\usepackage{color}
\usepackage{enumitem}
\usepackage{mathtools}
\usepackage{algpseudocode}
\usepackage{algorithmicx, algorithm}
\usepackage{booktabs}
\usepackage{subcaption}
\usepackage{xr}
\usepackage[T1]{fontenc}
\usepackage[utf8]{inputenc}

\numberwithin{equation}{section}
\newtheorem{theorem}{Theorem}[section]
\newtheorem{lemma}[theorem]{Lemma}

\newtheorem{remark}[theorem]{Remark}

\def\b{\beta}
\def\d{\delta}
\def\e{\epsilon}
\def\g{\gamma}

\def\s{\sigma}

\def\k{\kappa}
\def\lam{\lambda}
\def\L{\Lambda}

\def\D{\Delta}
\def\P{\Phi}

\def\bw{\mathbf w}

\def\by{\mathbf y}
\def\bx{\mathbf x}

\def\bg{\mathbf g}
\def\bW{\mathbf W}

\def\bB{\mathbf B}
\def\R{\mathbb R}
\def\P{\mathbb P}
\def\E{\mathbb E}

\def\bssigma{\boldsymbol{\sigma}}

\def\l{\left}
\def\r{\right}

\def\ll{\left\lVert}
\def\rl{\right\rVert}
\def\lv{\left\lvert}
\def\rv{\right\rvert}
\def\({\left(}
\def\){\right)}
\def\[{\left[}
\def\]{\right]}

\def\pt{\partial}
\def\nb{\nabla}

\def\qd{\quad}

\def\h{\hat}

\def\hp{\hat{p}}
\def\pinf{p_\infty}

\def\hd{\h{\d}}

\usepackage{xr}
\begin{document}
\title{Towards Theoretical Understanding of Large Batch Training  in Stochastic Gradient Descent}


%

\author{
  Xiaowu~Dai\thanks{Equal contribution.}  \thanks{Statistics, University of Wisconsin-Madison} \ and \ Yuhua~Zhu\footnotemark[1] \thanks{Mathematics, University of Wisconsin-Madison (Corresponding E-mail: \href{mailto:yzhu232@wisc.edu}{yzhu232@wisc.edu}).}}

  \date{}
  \maketitle

\begin{abstract}
Stochastic gradient descent (SGD)  is almost ubiquitously used for training non-convex optimization tasks. 
Recently, a hypothesis proposed by \citet{keskar} that \emph{large batch methods  tend to converge to sharp minimizers} has  received increasing attention. We theoretically justify this hypothesis by providing new properties of SGD in both finite-time and asymptotic regimes. In particular, we give an explicit escaping time of SGD from a local minimum in the finite-time regime and prove that SGD tends to converge to flatter minima in the asymptotic regime (although may take exponential time to converge) regardless of the batch size. 
We also find that SGD with a larger ratio of  learning rate to batch size 
 tends to converge to a flat minimum faster, however, its generalization performance
 could be worse than the SGD with a smaller ratio of  learning rate to  batch size.
We include experiments to corroborate these theoretical findings.
\end{abstract}

\noindent%

\section{Introduction}
\label{sec:intro}

Deep neural networks are typically trained by stochastic gradient descent (SGD) and  its variants. These methods update the weights using an estimated gradient from a small fraction of large training data. 
Although deep neural networks are highly complex and non-convex, the SGD training models possess good properties in the sense that saddle points can be avoided \citep{Ge2015} and ``bad" local minima vanish exponentially \citep{Choromanska2015, dauphin2014}.
However, a central challenge remains about why and when SGD training neural networks tend to generalize well to unseen data despite the fact of heavily over-parameterization and overfitting \citep{zhang2017}.

Recently, \citet{keskar} proposed a hypothesis based on empirical experiments that (i) large-batch methods tend to converge to sharp minimizers of the training function and (ii) the sharp minimum causes a worse generalization.  These two parts of the hypothesis are important for understanding the SGD in the deep neural networks. In this paper, we focus on the first part of the hypothesis.
Extensive numerical results corroborate the positive correlation between large-batch methods and sharp minimizers; see, e.g., \citet{Dinh, hoffer}. However, the theoretical result for supporting this observation is limited in the literature. Our work fills some gap in this important direction by 
providing new results on the properties of SGD in both \emph{finite-time} regime where the number of SGD iterations is finite and \emph{asymptotic} regime where the number of SGD iterations is sufficiently large. As a result, we can 
justify and provide new insights into the first part of the hypothesis by  \citet{keskar}.



 The main contributions of this paper are summarized as follows:
 \begin{itemize}
\item  We manage to use the finite-time escaping time of SGD from one local minimum to its nearest local minimum as an approach for justifying the hypothesis by  \citet{keskar}.
\item We prove that SGD tends to converge to flatter minima in the asymptotic regime regardless of the batch size. However, it may take exponential time to converge. This result provides new insights into  the hypothesis by  \citet{keskar}. 
\item We derive new results showing that the SGD with a larger learning rate to batch size ratio
 tends to converge to a flat minimum faster, however, its generalization performance
 could be worse than the SGD with a smaller  learning rate to batch size ratio.
\end{itemize}
\section{Main results}
\label{sec:mainresults}
Suppose the training set consists of $N$ samples. 
we define $L_n(\cdot)$ as the loss function for the sample $n\in\{1,\ldots,N\}$. 
Then $L(\cdot) = \E[L_n(\cdot)]$ is the risk function, where the expectation is taken with respect to the population of data.
Let $\bw$ be the vector of unknown model parameters in $\R^d$.


The mini-batch SGD estimates the gradient $\bg$ with some mini-batch $B$,  a set of $M$ randomly selected sample indices from $\{1,\ldots,N\}$, by
$\widehat{\bg}^{(B)}(\bw)=\frac{1}{M}\sum_{n\in B}\nabla L_n(\bw).$
We consider the stochastic gradient descent with learning rate $\gamma_k$  and mini-batch batch size $M_k$, and it gives the update rule
\begin{equation}
\label{eqn:minibatchsgd}
\bw_{k+1}  = \bw_k - \frac{\g_k}{M_k}\sum_{n\in B_k} \nabla L_n(\bw_k).
\end{equation}
Here, $k$ indexes the update step, and $|B_k| = M_k$.
We call (\ref{eqn:minibatchsgd}) a small batch training if $M_k<< N$ and typically  $M_k \in\{64, 128, 256\}$. In contrast, we call (\ref{eqn:minibatchsgd}) a \emph{large batch training} if $M_k/N$ is some non-negligible positive constant and typically $M_k/N=10\%$.

We allow the diminishing learning rate $\gamma_k$ and varying batch size $M_k$ in (\ref{eqn:minibatchsgd}), which is motivated  from  practice that SGD converges to the optimum by decreasing the learning rate. 

\begin{figure}
    \centering
    \includegraphics[width=\textwidth]{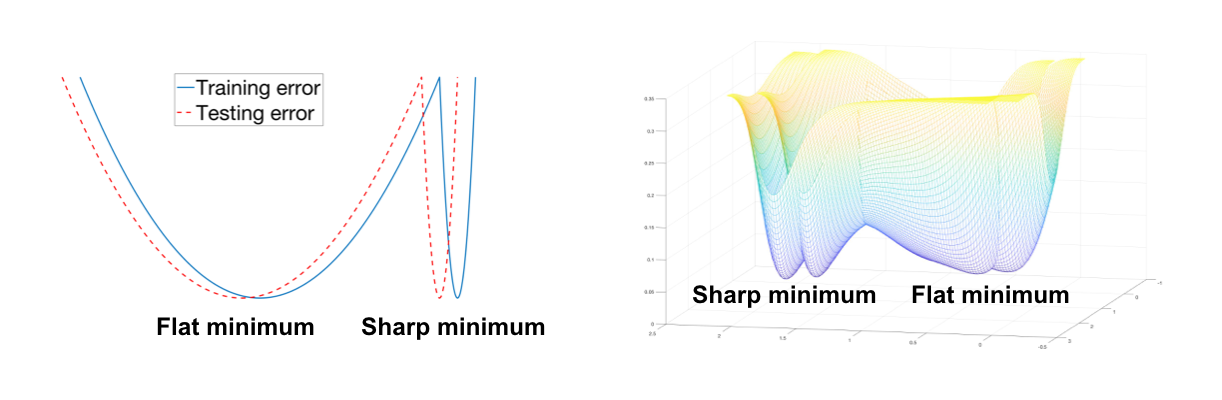}
    \caption{A  sketch of ``flat" and ``sharp" minima for one-dimensional case (left panel) and two-dimensional case (right panel). The vertical axis indicates the value of the loss function.}
    \label{fig:conceptualflatsharp}
\end{figure}

\paragraph{KMNST hypothesis}
\label{sec:drawbacks}
We call the following hypothesis proposed by \citet{keskar} as the KMNST hypothesis since K-M-N-S-T is the collection of author initials in \citet{keskar}:
 \begin{equation*}
 \label{eqn:kmnsthypo}
\text{ \emph{Large batch training tends to converge to the sharp minimizer of the training function}.}
 \end{equation*}
 A conceptual sketch of ``sharp" (and relatively, ``flat") minima is plotted in Figure \ref{fig:conceptualflatsharp}.  
The theory built in this Section \ref{sec:mainresults} aims to  justify the KMNST hypothesis.



\subsection{Stochastic differential equation for SGD}
\label{sec:sgdminibatch}

We consider SGD as a discretization of  stochastic differential equations. 
Let $\text{Var}[\nabla L_n(\bw)]\equiv\bssigma^2(\bw)$, which is finite and positive definite for typical loss functions. 
In Appendix \ref{sec:proofofexpvarlnbw}, we show that for independent and identically distributed (iid)  samples and $\bw$ in any bounded domain,
\begin{equation}
\label{eqn:expvarlnbw}
\E[\widehat{\bg}^{(B)}(\bw)] = \nabla L(\bw), \ \  \ \ \text{Var}[\widehat{\bg}^{(B)}(\bw)] = M^{-1}\bssigma^2(\bw).
\footnote{\cite{hoffer, Jastrzebski2017}  obtain a similar result as the (\ref{eqn:expvarlnbw}) but in  a different sense. 
Specifically, (\ref{eqn:expvarlnbw}) takes the expectation and variance with respect to the underlying population, however, 
\cite{hoffer, Jastrzebski2017} take the expectation and variance with respect to the sampling distribution of $B\in\{1,\ldots,N\}$. Note that (\ref{eqn:expvarlnbw}) is preferable if we want to analyze the risk function $L(\cdot)$ instead of the sample average loss $N^{-1}[L_1(\cdot)+\cdots+L_N(\cdot)]$ and if we regard the training data only a subset of the true underlying population. }
\end{equation}
We can write the mini-batch SGD (\ref{eqn:minibatchsgd}) as
\begin{equation*}
\bw_{k+1}  = \bw_k - \gamma_k\nabla L(\bw_k) + \frac{\gamma_k}{\sqrt{M_k}}\boldsymbol{\epsilon},
\end{equation*}
where $\boldsymbol{\epsilon}$ has zero mean and variance $\bssigma(\bw)$ by (\ref{eqn:expvarlnbw}). We consider a stochastic differential equation (SDE):
\begin{equation}
\label{eqn:sde}
d\bW(t) = -\nabla L(\bW(t))dt - \sqrt{\frac{\gamma(t)}{M(t)}}\bssigma(\bW(t))d\bB(t), \ \ \bW(0) = \bw_0.
\end{equation}
By the Euler scheme, the SDE (\ref{eqn:sde}) can be discretized  to obtain the mini-batch SGD (\ref{eqn:minibatchsgd}); see, e.g., \cite{mandt2017, Jastrzebski2017, li2017}. 
The stochastic Brownian term $\bB(t)$ in (\ref{eqn:sde}) accounts for the random fluctuations due to the use of mini-batches for gradient estimation in (\ref{eqn:minibatchsgd}). 
Note that (\ref{eqn:sde}) allows the batch size and step size to be time-dependent.

We consider the gradient covariance to be isotropic: 
\begin{equation}
\label{eqn:isotconv}
\bssigma^2(\bw)= \beta(\bw)\cdot\mathbf{I},
\end{equation} where $\beta(\bw)$ may depend on $\bw$. 
A similar assumption has been made in the literature, see e.g.,  \citet{Jastrzebski2017, Chaudhari2017deep},  where they assume $\beta(\bw) \equiv \beta$ is a constant. 
Let $p(\bw,t)$ be the probability density function of the solution $\bW(t)$  to the SDE  (\ref{eqn:sde}).
We derive the following characteristics for $p(\bw,t)$ in Appendix \ref{sec:proofoflemfokkerplank}. 
\begin{lemma}
\label{lem:fokkerplank}
The $p(\bw,t)$ satisfies the following  Fokker-Planck equation:
\begin{equation}
\label{eqn:probdensitypthetatw}
\partial_tp = \nabla\cdot\left(\left[\nabla \left(L(\bw)+ \frac{\gamma(t)\beta(\bw)}{2M(t)}\right)\right]p +  \frac{\gamma(t)\beta(\bw)}{2M(t)}\nabla p\right), \quad p(\bw,0)=\delta(\bw_0),
\end{equation}
where  $\delta(\cdot)$ denotes the delta function.
\end{lemma}
Note that the drift term in (\ref{eqn:probdensitypthetatw}) is $-\nabla [L(\bw)+ \gamma(t)\beta(\bw)/\{2M(t)\}] \neq -\nabla L(\bw)$, which implies the SGD does not follow the mean drift $-\nabla L(\bw)$ to be its update direction.
Specifically, a larger $\gamma(t)/M(t)$ ratio corresponds to a drift term deviate more from the mean drift $-\nabla L(\bw)$. This sheds light on the possible case that even 
the SGD with a larger $\gamma(t)/M(t)$ ratio tends to converge to a flat minimum faster (to be justified in Section \ref{sec:proofofhypothesis}), its generalization performance
 could be worse than the SGD with a smaller $\gamma(t)/M(t)$ ratio (to be illustrated in Section \ref{sec:simulation}).

The results derived in this Section \ref{sec:sgdminibatch} can be related with the KMNST hypothesis in the following sense: the dynamics of SGD would depend on the $\gamma(t)/M(t)$ ratio instead of the $\gamma$ or $M$ separately, which is clear from the experiments in Section \ref{sec:simulation}.

\subsection{KMNST hypothesis in the finite-time regime}
\label{sec:finitetime}

\begin{figure}
    \centering
    \includegraphics[width=0.5\textwidth]{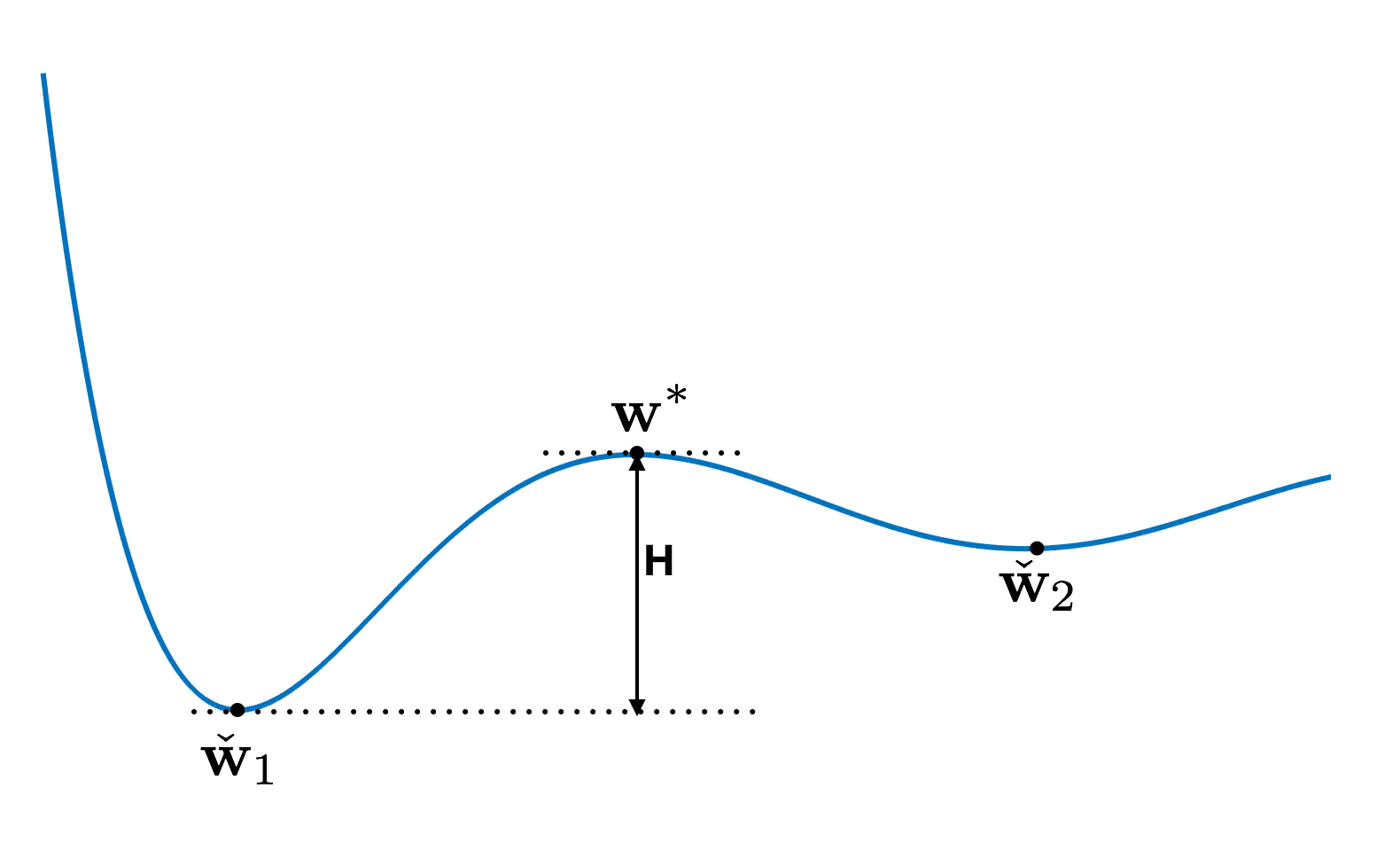}
    \caption{A  sketch of two local minimizer $\check{\bw}_1$ and $\check{\bw}_2$ of a risk function. The $\bw^*$ is the saddle point between $\check{\bw}_1$ and $\check{\bw}_2$ and the $H$ is the relative height of $\bw^*$ to $\check{\bw}_1$.}
    \label{fig:finite_time_pic}
\end{figure}

We first consider the behavior of SGD in the finite-time regime $t<\infty$, which is typical in the practice.
Specifically, we are interested in the escape time of SGD from one local minimizer $\check{\bw}_1$ to its   nearest local minimizer $\check{\bw}_2$. Refer to the Figure \ref{fig:finite_time_pic} as an illustration. 
 Let $\bw^*$ be the saddle point\footnote{There are possibly multiple saddle points between $\check{\bw}_1$ and $\check{\bw}_2$. We define $\bw^*$ as the saddle point with the minimal height among all saddle points in the following sense. Let $w(t),0\leq t\leq 1,$ be any continuous path from $\check{\bw}_1$ to $\check{\bw}_2$. Denote by $\widehat{w} = \arg{\text{inf}}_{w:w(0)=\check{\bw}_1, w(1) = \check{\bw}_2}\sup_{t\in[0,1]}L(w(t))$ the path with the minimal saddle point height among all continuous path. We define that $\bw^* = \max_{t\in[0,1]}\widehat{w}(t)$.}  between $\check{\bw}_1$ and $\check{\bw}_2$.  By the definition of $\bw^*$,  the Hessian $\D  L(\bw^*)$ can be shown to have a single negative eigenvalue $-\lam^*$ (e.g., \citet{berglund2013kramers}). 
 By the Eyring-Kramers formula, we have the following theorem.
\begin{theorem}
\label{thm: exitingtime}
Let $\tau_{\check{\bw}_1 \to \check{\bw}_2}$ be the transition time from $\check{\bw}_1$ to $\check{\bw}_2$ for $\bW(t)$, then
\begin{equation*}
\mathbb{E}[\tau_{\check{\bw}_1 \to \check{\bw}_2}] = \frac{2\pi}{\lam^*} \sqrt{\frac{|\Delta L(\bw^*)|}{|\Delta L(\check{\bf{w}}_1)|}}e^{H \cdot 2M(\check{\bw}_1)/[\gamma(\check{\bw}_1)\b(\check{\bw}_1)]} \{1+O\(\sqrt{\e}\log(\e^{-1})\)\} \label{eq: extingtime}
\end{equation*}
where $|\D L(\cdot)|$ represents for the determinate of the Hessian,  $H = H(\bw^*,\check{\bw}_1) \equiv L(\bw^*) - L(\check{\bw}_1)$  is the relative height of $\bw^*$ to $\check{\bw}_1$, $M(\check{\bw}_1)$ is the batch size of the SGD at $\check{\bw}_1$, $\gamma(\check{\bw}_1)$ is the learning rate of the SGD  at $\check{\bw}_1$, and $\beta$ is defined in (\ref{eqn:isotconv}).
\end{theorem}

The above theorem is proved  by \citet{bovier2004metastability, bovier2005metastability} and in a more general case by \citet{berglund2013kramers}. From this theorem, one can see that the transition time depends on three factors, the diffusion factor $\gamma\beta/M$ in the SGD, the potential barrier $H(\bw^*,\check{\bw}_1)$ that SGD has to climb in order to escape $\check{\bw}_1$, and the determinant of the Hessian at $\check{\bw}_1$ and $\bw^*$. 

The results shown in this Section \ref{sec:finitetime} can explain the KMNST hypothesis as follows.  A larger batch size $M$ of SGD at  local minimizer $\check{\bw}_1$ corresponds to a longer escaping time from $\check{\bw}_1$, which is modeled by $\mathbb{E}[\tau_{\check{\bw}_1 \to \check{\bw}_2}]$. 
Hence, even if $\check{\bw}_1$ corresponds to a sharp minimum with a large $|\Delta L(\check{\bf{w}}_1)|$, the exponential term $\exp[H \cdot 2M(\check{\bw}_1)/[\gamma(\check{\bw}_1)\b(\check{\bw}_1)]]$ could dominate the escaping time. As a result, the large batch training will be trapped at a sharp minimizer in the finite-time regime, which is the same as observed by \citet{keskar} that  \emph{large batch training tends to converge to the sharp minimizer of the training function}.
On the other hand, if the batch size is small, then $\exp[H \cdot 2M(\check{\bw}_1)/[\gamma(\check{\bw}_1)\b(\check{\bw}_1)]]$ is small. As a result, only when $|\D L(\check{\bw}_1)|$ is small enough, then SGD can be trapped at this minimizer, which implies that small batch training tends to converge to flatter minima. 

However, these phenomena will change in the asymptotic regime $t\to\infty$ as explained in Section \ref{sec:proofofhypothesis}.

\subsection{KMNST hypothesis in the asymptotic regime}
\label{sec:proofofhypothesis}

\paragraph{Main assumptions} In this section, we consider  the asymptotic regime that $t\to\infty$ and suppose the following three  assumptions\footnote{We note that if the parameter vector $\bw$ lies in a bounded region, then the Gibbs density is well defined only if $\int e^{-L(\bw)}d\bw<\infty$, the Poincar\'e inequality is always true, and the assumption (A.3) is always true. Thus, although the mean cross entropy loss with bounded parameters does not satisfy (A.1) or (A.2), our results in this section still hold for the mean cross entropy loss.}:
\begin{itemize}
\item[(A.1)] $L(\bw)$ is  {\emph{confinement}}: $\lim_{\|\bw\|\to +\infty}L(\bw) = +\infty$ and $\int e^{-L(\bw)} d\bw < +\infty$.
\item[(A.2)]  $\lim_{\|\bw\|\to +\infty}  \left\{\|\nb L(\bw)\|^2/2 - \nb\cdot\nb L(\bw)\right\} = + \infty$, where $\nb\cdot\nb L$ denotes the trace of the Hessian for $L$. Moreover, $\lim_{\|\bw\|\to +\infty}  \left\{\nb\cdot\nb L(\bw)/\|\nb L(\bw)\|^2\right\} = 0$.
\item[(A.3)]  There exists a constant $M$, such that $\lv e^{- L(\bw)}\(\ll \nb L(\bw) \rl^2 - \nb\cdot\nb L(\bw)\) \rv \leq M$.
\end{itemize}
We show in Appendix \ref{sec:disonassump} that (A.1) -- (A.3) hold for typical loss functions such as the regularized mean cross entropy and the square loss functions.
These assumptions  appear commonly in the diffusion process literature, see, e.g., \citet{pavliotis2014stochastic}.
In particular, (A.1) ensures the  Gibbs  density function $p_G(\bw) =  e^{-L(\bw)}$ is well defined, and (A.2) is sufficient for the measure $\mu(\bw) = \int p_G(\bw)d\bw =  \int e^{-L(\bw)}d\bw$ to satisfy the Poincar\'e inequality (e.g., \citet{pavliotis2014stochastic, raginsky2017non}):
\begin{equation}
\label{eqn:poincare}
\begin{aligned}
  \int \ll \nb f(\bw) \rl^2d\mu(\bw)  \geq C_P\int \(f(\bw) - \int f(\bw) d\mu(\bw)\)^2 d\mu(\bw), \text{ for some } C_P>0,
  \end{aligned}
\end{equation}
holds for any $f$ satisfying $\int f^2(\bw)d\bw<\infty$.

We first give the stationary solution for the Fokker-Planck equation (\ref{eqn:probdensitypthetatw}) when $t\to\infty$.
\begin{lemma}
\label{lem:stationbeta}
Under the assumption (A.1) and suppose $\beta(\bw)\equiv\beta$, then (\ref{eqn:probdensitypthetatw}) has a stationary solution
\begin{equation*}
\label{eqn:defofpinf}
p_\infty(\bw) = \kappa e^{-\eta_\infty L(\bw)}, 
\end{equation*}
where 
\begin{equation*}
\eta_\infty  = 2M/[\gamma\b(\check{\bw})]
\end{equation*} 
with the limiting batch size $M=\lim_{t\to\infty}M(t)$, the limiting learning rate $\gamma=\lim_{t\to\infty}\gamma(t)$, and the convergent local minimizer $\check{\bw}$.
The constant $\kappa$ in the above formula is a normalization factor such that $\int p_\infty(\bw)=1$. 
\end{lemma}
Proof for this lemma  is given in Appendix \ref{sec:proflemstationbeta}.
We remark that for general $\beta(\bw)$ depending on $\bw$, the existence and an explicit form of  stationary solution for  (\ref{eqn:probdensitypthetatw}) remain an open question in the literature.
Hence, we focus on $\beta(\bw)\equiv\beta$ in this section.

Similar results as Lemma \ref{lem:stationbeta} can be found  in related work, e.g., \citet{Jastrzebski2017}. 
However, it is not  clear whether  $p(\bw,t)$ converges to $p_{\infty}(\bw)$, not to mention how fast that $p(\bw,t)$ would converge to $p_{\infty}(\bw)$. 
The following theorem gives a positive answer to this question, which later provides a new insight into the justification of KMNST hypothesis.
\begin{theorem}
\label{thm:sigm4.1proof}
Under assumptions (A.1) -- (A.3), the probability density function $p(\bw,t)$  of $\bW(t)$    converges to the stationary solution $\pinf(\bw)$.
Moreover, there exists $T>0$ such that for any $t>T$,  
\begin{equation*}
 \ll \frac{p(\bw,t) - p_\infty(\bw)}{\sqrt{p_\infty(\bw)}} \rl_{L^2(\R^d)}^2 \leq C(t,T)e^{-C_P\cdot (t-T)/\eta_\infty},
\end{equation*}
where $C_P$ is a constant  define in (\ref{eqn:poincare}) and $C(t,T) =C_P\cdot (t-T)/\eta_\infty +  \ll({p(T) - p_\infty})/\sqrt{p_\infty} \rl_{L^2(\R^d)}^2$. 
\end{theorem}
The proof for this theorem is given in Appendix \ref{sec:proofofthm:sigm4.1proof}. We also give a quantification of constant $T$ in Appendix \ref{sec:quantificT}.
Three remarks on Theorem \ref{thm:sigm4.1proof} are as follows.
\emph{First}, this theorem shows that $p(\bw,t)$ always converges to $p_\infty$ with an exponential rate regardless of the initial value. This theorem provides a theoretical ground for the work that manages to understand $p(\bw,t)$ based on analysis of the stationary distribution $p_\infty$ (see,  e.g., \citet{Jastrzebski2017}).
\emph{Second}, it is known \citep{raginsky2017non} the Poincar\'e constant $C_P\propto e^d$, where $d$ is the dimension of the parameter $\bw$. In the setting of the deep neural networks, $C_P$ can be very large and it takes exponential time $t>e^{d}$ such that $p(\bw,t)$ would approach to the stationary distribution $p_\infty$. 
Therefore, the results only based on the stationary solution do not reveal information in the finite-time regime.
\emph{Third}, the convergence rate is relatively faster with a larger $\gamma/M$ since it corresponds to a smaller $\eta_\infty$. The last two remarks are illustrated by experiments in Section \ref{sec:simulation}.

We now characterize $\bW(t)$ in the asymptotic regime  $t\to\infty$ based on the stationary distribution $p_\infty$, and we give the proof of the following theorem in Appendix \ref{eqn:proofofthfinal}.
\begin{theorem}
\label{thm:probmainresult4111}
Let $\check{\bw}$ be a  local minimizer. Then,
\begin{equation*}
\begin{aligned}
\lim_{\e\to0}\P(|\bW(\infty) - \check{\bw}|\leq \epsilon)  =\frac{\kappa e^{-2\eta_\infty L(\check{\bw})}}{\eta_\infty^{d/2}\sqrt{|\mathrm{\Delta}L(\check{\bw})|}}\lim_{\e\to0}\l[e^{\eta_\infty\epsilon^2}\prod_{j=1}^d\sqrt{1-e^{-\e^2\eta_\infty\lam_j/\pi}}\r],
\end{aligned}
\end{equation*}
where  $d$ is the dimension of $\bw$, $\lambda_j$s and $|\mathrm{\Delta}L(\check{\bw})|$ represent  the eigenvalues and the determinant of  loss function Hessian  $\mathrm{\Delta}L(\check{\bw})$, respectively, and the constants $\kappa, \eta_\infty$ are defined  in Lemma \ref{lem:stationbeta}.
\end{theorem}
Given the complex form of the probability in Theorem \ref{thm:probmainresult4111}, we give numerical illustrations in Appendix \ref{sec:numerprob}.
To appreciate the implication of Theorem \ref{thm:probmainresult4111}, we consider any two local minimizers 
$\check{\bw}_1$ and $\check{\bw}_2$ with  the same value of $L(\check{\bw}_1)=L(\check{\bw}_2)$.
Then, 
\begin{equation}
\label{eqn:ratioofprob}
\begin{aligned}
&\lim_{\e\to0}\frac{\P(|\bW(\infty) - \check{\bw}_1|\leq \epsilon) }{\P(|\bW(\infty) - \check{\bw}_2|\leq \epsilon)} = \sqrt{\frac{\lv\D L(\check{\bw}_2)\rv}{\lv\D L(\check{\bw}_1)\rv}}.
\end{aligned}
\end{equation}
The ratio of probability (\ref{eqn:ratioofprob}) 
implies that in the asymptotic regime $t\to\infty$, the probability of SGD converging to a flatter minimum with a smaller determinant $|\D L(\cdot)|$ is always larger than the probability of SGD converging to a sharper minimum with a larger determinant $|\D L(\cdot)|$.
Moreover, (\ref{eqn:ratioofprob}) does not depend on the batch size or learning rate, but it only depends on the determinant of Hessian at the local minimum.

The results derived in this Section \ref{sec:proofofhypothesis} provide some new insights into the KMNST hypothesis:  SGD tends to converge to flatter minima regardless of the batch size $M$ (or the ratio $\gamma/M$) in the asymptotic regime $t\to\infty$ as shown by (\ref{eqn:ratioofprob}). However, it may take exponential time $e^d$ to converge, where $d$ is the dimension of the model parameter $\bw$. The experiments in Section \ref{sec:simulation} further corroborate these theoretical finding.

\section{Numerical Experiments}
\label{sec:simulation}

In this section, we show experiments to corroborate the theoretical findings in the previous section. We train 4-layer batch-normalized ReLU MLPs on MNIST with different learning rate $\gamma$ and batch size $M$.
Specifically, we use three $\gamma/M$ ratios:  $\gamma/M = 0.01/128, 0.1/128, 0.2/256$. 
As is common for such tasks, the mean cross entropy loss is used as the loss function. We discussed in Section \ref{sec:sgdminibatch} that this loss satisfies our assumptions for theoretical analysis.

\begin{figure}
    \centering
    \includegraphics[width=\textwidth]{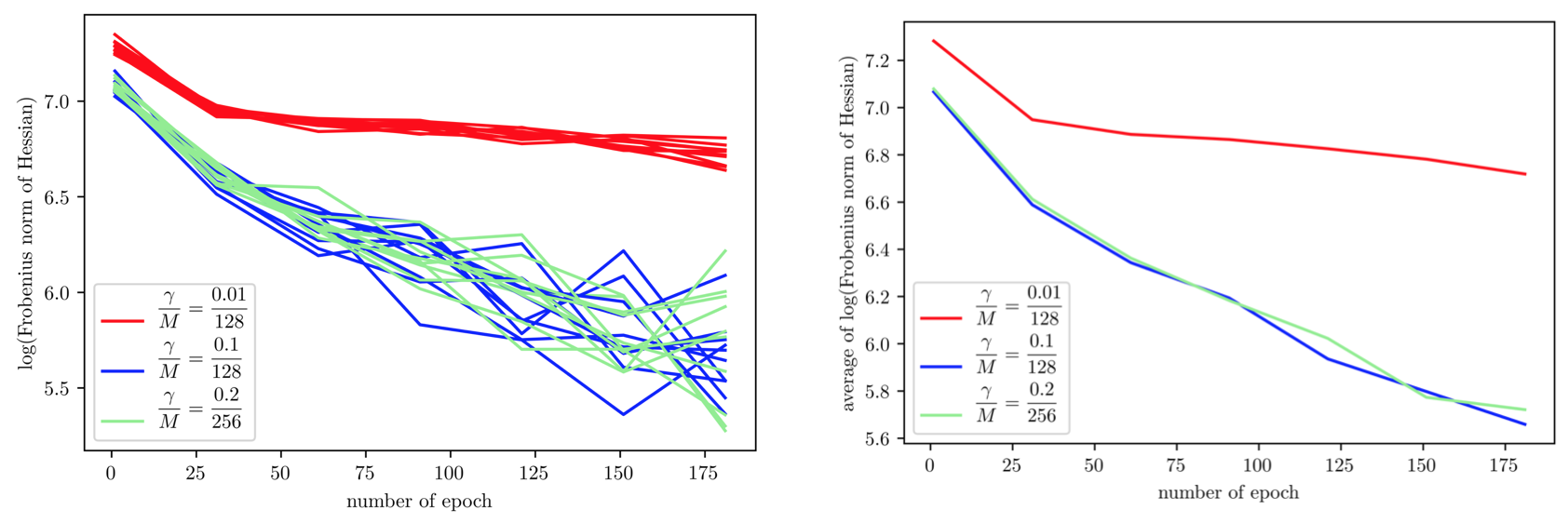}
    \caption{Log of Frobenius norm of Hessian as a function of epochs. Three $(\gamma,M)$ pairs  $(0.01, 128)$, $(0.1, 128)$ and $(0.2, 256)$ are studied, which are denoted in red, blue and green,  respectively. The left plot shows 10 experiments for each of three $(\gamma,M)$ pairs and the right plot shows the average of 10 experiments. Total 180 epochs are trained.}
    \label{fig:hessian-two}
\end{figure}

\paragraph{Geometry of SGD updates}
Figure \ref{fig:hessian-two} shows the log of Frobenius norm of Hessian for minima obtained by SGD. 
Due to the high computational cost for computing the determinant of the Hessian, we use the 
Frobenius norm of the Hessian as a substitute.  Note that a larger Frobenius norm of Hessian corresponds to a sharper minimum. The Frobenius norm is approximated using the method in \citet{wu2017}. 
Note that the dynamics of SGD behave similar across 10 experiments for each of three $\gamma/M$ ratios as shown in the left plot of Figure \ref{fig:hessian-two}. Hence we focus on the averaged dynamics as in the right plot of Figure \ref{fig:hessian-two}.
Three main results can be observed from Figure \ref{fig:hessian-two}:
\begin{itemize}
\item First,  for the same $\gamma/M$ ratio (e.g., $\gamma/M = 0.1/128$ and $0.2/256$), the minima obtained by SGD have the very similar norm of the Hessian. This illustrates the Lemma \ref{lem:fokkerplank}, \ref{lem:stationbeta} and Theorem \ref{thm: exitingtime} that the dynamics and geometry of the minima obtained by SGD would depend on the ratio $\gamma/M$ instead of individual $\gamma$ or $M$ separately. A similar phenomenon is also observed by \citet{Jastrzebski2017}.
\item Second, since the SGD is trained using 180 epochs, the dynamics of SGD in Figure \ref{fig:hessian-two} fall in the finite-time regime.
It is clear that the rate of SGD tending to a flatter minimum (i.e., with a smaller norm of the Hessian) with a larger $\gamma/M$  ratio (e.g., $\gamma/M = 0.1/128$)  is faster compared to with a smaller $\gamma/M$ ratio (e.g., $\gamma/M = 0.01/128$). This illustrates the finite-time analysis in 
Theorem \ref{thm: exitingtime} that the SGD with a smaller $\gamma/M$ ratio is easier to be trapped around a minimum and hence the SGD tends to other minima slower. As a result, the Hessian of minima changes slower for SGD with a smaller $\gamma/M$ ratio.
\item Third, Figure \ref{fig:hessian-two}  also sheds light on the dynamics of SGD in the asymptotic regime. The SGD tends to converge to a flatter minimum regardless of the $\gamma/M$ ratio, which demonstrates Theorem \ref{thm:probmainresult4111} and its corollary (\ref{eqn:ratioofprob}). However, the convergence rate is slow, in particular for the SGD with a small $\gamma/M$ ratio, which is theoretically shown in Theorem \ref{thm:sigm4.1proof} and its following remarks.
\end{itemize}

\paragraph{Training and generalization of SGD}

\begin{figure}
    \centering
    \includegraphics[width=\textwidth]{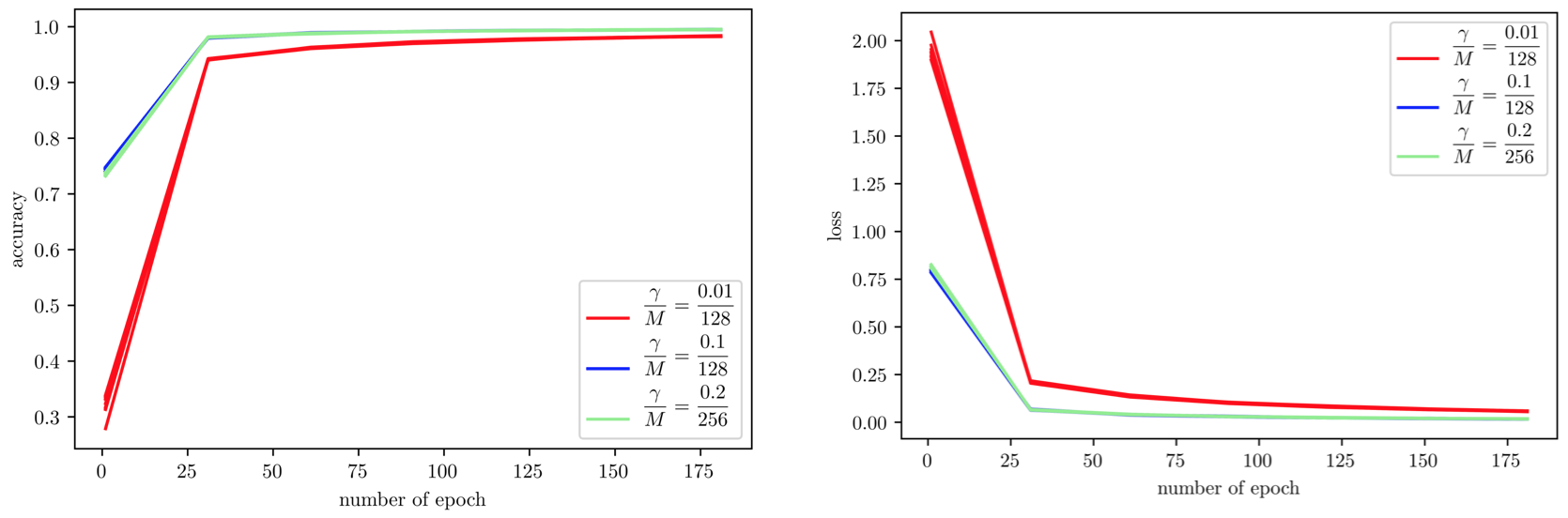}
    \caption{The left plot shows the training accuracy as a function of epochs and the right plot shows the cross entropy loss as a function of epochs.
    Three $(\gamma,M)$ pairs $(0.01, 128)$, $(0.1, 128)$ and $(0.2, 256)$ are studied, which are denoted in red, blue and green,  respectively. Both plots show 10 experiments for each of three $(\gamma,M)$ pairs. Total 180 epochs are trained.}
    \label{fig:training-loss}
\end{figure}

\begin{figure}
    \centering
    \includegraphics[width=\textwidth]{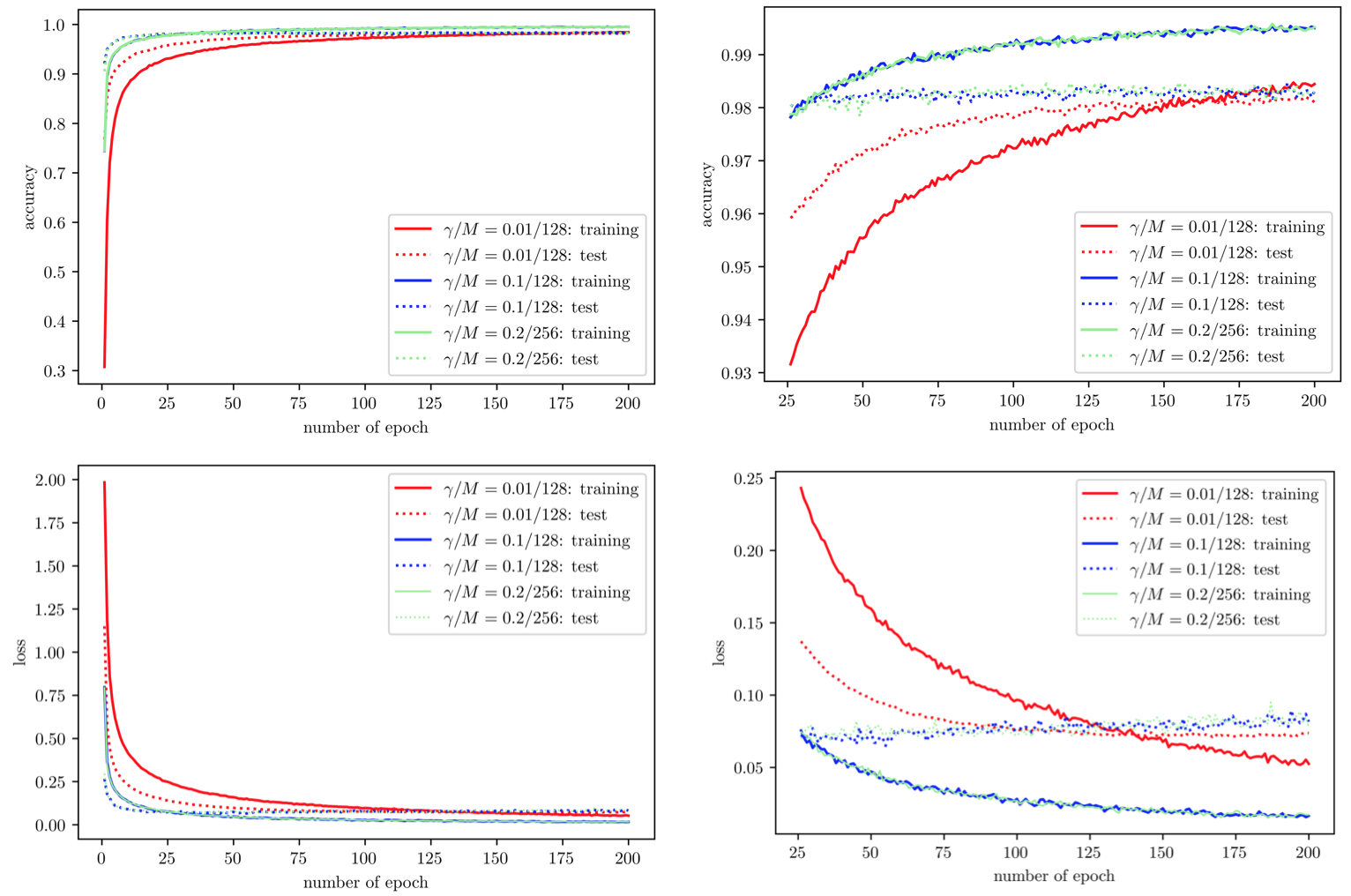}
    \caption{The top left plot shows the training and test accuracy as a function of epochs. The top right plot gives the zoomed in performance of the accuracy when epochs are no less than 25. 
    The bottom left plot shows the cross entropy loss as a function of epochs. The bottom right plot gives the zoomed in performance of the loss when epochs are no less than 25. 
    Three $(\gamma,M)$ pairs $(0.01, 128)$, $(0.1, 128)$ and $(0.2, 256)$ are studied, which are denoted in red, blue and green,  respectively.  Total 200 epochs are trained.}
    \label{fig:loss-accuracy-once}
\end{figure}

Figure \ref{fig:training-loss} shows the training accuracy and loss for the model trained by SGD.
We run 10 experiments. It is clear that the  training accuracy and loss are very close across 10 experiments
 for each of three $\gamma/M$ ratios.  Thus, we focus on interpreting the training and generalization performance of the model obtain from one experiment, which is shown in Figure \ref{fig:loss-accuracy-once}.
Three main results can be observed from Figure \ref{fig:loss-accuracy-once}:
\begin{itemize}
\item First,  for the same $\gamma/M$ ratio (e.g., $\gamma/M = 0.1/128$ and $0.2/256$), the training error and test error are very close. This meets our expectation since the dynamics of SGD only depends on the ratio $\gamma/M$  as discussed above and the models trained by SGD with the same $\gamma/M$ ratio  should behave similarly.
\item Second, the model obtained with a larger $\gamma/M$ ratio (e.g., $\gamma/M = 0.1/128$) gives a better training accuracy and a smaller training loss compared with the case of a smaller $\gamma/M$ ratio (e.g., $\gamma/M = 0.01/128$). This can be partially justified by our finite-time analysis in
Theorem \ref{thm: exitingtime} that the SGD with a larger $\gamma/M$ ratio is easier to escape a local minimum.
\item  Third, the model obtained with a smaller $\gamma/M$ ratio gives a \emph{smaller} test loss after a certain time (it is after 100 epochs in the bottom right plot of Figure \ref{fig:loss-accuracy-once}). This can be explained by Lemma \ref{lem:fokkerplank} and its following remark. Specifically,   a smaller $\gamma/M$ ratio corresponds a mean drift deviates less from the mean drift $-\nabla L(\bw)$, where $-\nabla L(\bw)$ is the drift for a global minimizer of the risk function $L(\bw)$. This shows a tradeoff between the large and small $\gamma/M$ ratio in the sense of the training and test loss.
\end{itemize}


\section{Related work}

The modeling of SDE for approximating SGD is well studied in the literature. See, e.g., \citet{mandt2017, poggio2017theory, li2017, Jastrzebski2017, chaudhari2017stochastic} and the references therein. Different from these work, we give a new result in Lemma \ref{lem:fokkerplank}, which not only gives the dynamics of SDE solution but also connects with the generalization performance. 
We also derive the theory for the SDE solution in the asymptotic regime, especially the convergence rate. 

We clarify the definition of the sharpness in multi-dimensional cases. We find that the production of eigenvalues, or equivalently, the determinant of  the risk function Hessian at minimizers  is appropriate. Similar results have been derived in \citet{Jastrzebski2017, dziugaite2017computing}.

The work by \citet{Jastrzebski2017} remarkably emphasize how the learning rate to batch size ratio affects the SGD and they also relate with the KMNST hypothesis. Here are some differences between \citet{Jastrzebski2017} and ours.
\begin{itemize}
\item \citet{Jastrzebski2017} use the stationary probability $p_\infty(\bw)$  to explain that the behavior of the SGD. However, we show that it takes the exponential time for $p(\bw,t)$ to converge to $p_\infty(\bw)$ in the setting of the deep neural network. Hence, $p_\infty(\bw)$ cannot fully explain the behavior of SGD in the practical  finite-time regime.  Our work adds new elements to this picture by studying the escaping time of SGD from a local minimum in the sense of finite-time regime and we also give a new result on the convergence rate of $p(\bw,t)\to p_\infty(\bw)$.
\item In particular, the stationary probability in \citet{Jastrzebski2017}  can not explain the KMNST hypothesis when two local minimizers $\check{\bw}_1$ and $\check{\bw}_2$ having a same risk $L(\check{\bw}_1) = L(\check{\bw}_2)$. In this case, the result of \citet{Jastrzebski2017}  coincides with (\ref{eqn:ratioofprob})  and it  is independent of $M$ or $\gamma$, which is undesired in explaining the KMNST hypothesis. 
On the other hand, there may exist many minima with a same risk value but different Hessians for a deep neural network. Therefore, our finite-time results can give a better explanation to the KMNST hypothesis in this case.
\end{itemize}




\section{Conclusion}
\label{sec:discussion}

In this paper, we investigate the relationship between the sharpness of the minima that SGD converges to with the ratio of the step size and the batch size. Using the SDE as the approximation of SGD, we explain part of the hypothesis proposed by \citet{keskar} that large-batch methods tend to converge to sharp minimizers of the training function using the escaping time theorem in the finite-time regime. We prove that for the isotropic case the probability density function of SGD will converge to the stationary solution for any initial data regardless of the time varying step size and batch size. We give the convergence rate, which indicates that with a larger ratio of the learning rate and the batch size, the probability will converge faster to the stationary solution. Asymptotically the probability of converging to the global minimum is independent of the batch size and learning rate, but it only depends on the sharpness of the minimum. We verify these theoretical findings with numerical experiments. 

There are many directions for further study such as how the ratio of the step size and batch size influence the generalization error. In our experiment, it indicates that with a larger learning rate to batch size ratio the generalization error is worse. Further theoretical analysis is desired. Another interesting topic is to study
 the stationary solution and the evolution the probability density function of SGD 
when the variance matrix is anisotropic, which remain open questions.


\section*{Appendix}

\appendix
\section{Proof of (\ref{eqn:expvarlnbw})}
\label{sec:proofofexpvarlnbw}

For the first part, without less of generality, we consider $\bw$ is in any bounded domain of $\R$.  Then
\begin{equation*}
\begin{aligned}
\nabla L(\bw) =  \frac{d}{d\bw}\E[L_n(\bw)] 
&= \lim_{h\to0}\frac{1}{h}\left\{\E[L_n(\bw+h)] - \E[L_n(\bw)]\right\}\\
&= \lim_{h\to0}\E\left\{\frac{L_n(\bw+h) - L_n(\bw)}{h}\right\} = \lim_{h\to0}\E\left\{\nabla L_n(\bw+\tau(h))\right\}.
\end{aligned}
\end{equation*}
where the last step is by the mean value theorem with some $0< \tau(h)< h$. Due to continuity of $\nabla L_n$, we can use the dominated convergence theorem and have
\begin{equation*}
\lim_{h\to0}\E\left\{\nabla L_n(\bw+\tau(h))\right\} = \E\left\{\lim_{h\to0}\nabla L_n(\bw+\tau(h))\right\}  = \E\left\{\nabla L_n(\bw)\right\}.
\end{equation*}
This completes the proof of the first part.
By assuming the iid of the data, we have $\text{Var}[\widehat{\bg}^{(B)}]  = M^{-1}\bssigma^2(\bw)$. This completes the proof of the second part.

\section{Proof of Lemma \ref{lem:fokkerplank}}
\label{sec:proofoflemfokkerplank}

We start to consider when $\beta(\bw) \equiv\beta$ is a constant and follow the strategy in \cite{kolpas2007coarse} to derive the Fokker-Planck equation.
First, consider $\bW(t)=W(t)\in\R$. Note that for SGD the corresponding $W(t)$ is a Markov process,
then the Chapman-Kolmogorov equation gives
\begin{equation*}
p\left(W(t_3)|W(t_1)\right) = \int_{-\infty}^{+\infty} p\left(W(t_3)|W(t_2)=w\right)p\left(W(t_2)=w|W(t_1)\right)dw.
\end{equation*}
Consider the integral
\begin{equation*}
I = \int_{-\infty}^{+\infty}h(w)\partial_t p(w,t|W)dw,
\end{equation*}
where $h(w)$ is a smooth function with compact support. Observe that
\begin{equation*}
\int_{-\infty}^{+\infty}h(w)\partial_t p(w,t|W)dw = \lim_{\Delta t\to 0 }\int_{-\infty}^{+\infty}
h(w)\left(\frac{p(w,t+\Delta t|W) - p(w,t|W)}{\Delta t}\right)dw.
\end{equation*}
Letting $Z$ be an intermediate point. Applying the Chapman-Kolmogorov identity on the right hand side yields
\begin{equation*}
\lim_{\Delta t\to 0 }\frac{1}{\Delta t}\left(\int_{-\infty}^{+\infty}h(w)\int_{-\infty}^{+\infty}p(w,\Delta t|Z)p(Z,t|W)dZdw-\int_{-\infty}^{+\infty}h(w)p(w,t|W)dw\right).
\end{equation*}
By changing the limits of integration in the first term and letting $w$ approach $Z$ in the second term, we obtain
\begin{equation*}
\lim_{\Delta t\to 0}\frac{1}{\Delta t}\left(\int_{-\infty}^{+\infty}p(Z,t|W)\int_{-\infty}^{+\infty}p(w,\Delta t|Z)(h(w)-h(Z))dwdZ\right).
\end{equation*}
Expand $h(w)$ as a Taylor series about $Z$, we can write the above integral as
\begin{equation*}
\lim_{\Delta t\to 0}\frac{1}{\Delta t}\left(\int_{-\infty}^{+\infty}p(Z,t|W)\int_{-\infty}^{+\infty}p(w,\Delta t|Z)\sum_{n=1}^{\infty}h^{(n)}(Z)\frac{(w-Z)^n}{n!}\right)dwdZ.
\end{equation*}
Now we define the function 
\begin{equation*}
D^{(n)}(Z)=\frac{1}{n!}\frac{1}{\Delta t}\int_{-\infty}^{+\infty}p(w,\Delta t|Z)(w-Z)^ndw.
\end{equation*}
We can write the integral $I$ as
\begin{equation*}
\int_{-\infty}^{+\infty}h(w)\partial_t p(w,t|W)dw = \int_{-\infty}^{+\infty}p(Z,t|W)\sum_{n=1}^{\infty}D^{(n)}(Z)h^{(n)}(Z)dZ.
\end{equation*}
Integrating by parts $n$ times gives
\begin{equation*}
\partial_t p(w,t) = \sum_{n=1}^{\infty}-\frac{\partial^n}{\partial Z^n}\left[D^{(n)}(Z)p(Z,t|W)\right].
\end{equation*}
Let $D^{(1)}(w) = -L(w)$, 
$D^{(2)}(w) = -\gamma(t)\beta/[2M(t)]$ and $D^{(n)}(w) = 0$ for all $n\geq 3$, the above equation yields
\begin{equation*}
\partial_t p(w,t) =  \frac{\partial}{\partial w}\left[\nabla L(w)p(w,t)\right] + \frac{\partial}{\partial w^2}\left[\frac{\gamma(t)\beta}{2M(t)}p(w,t)\right],
\end{equation*}
which is the Fokker-Planck equation in one variable. 
For the multidimensional case $\bW=(W_1,W_2,\ldots,W_p)\in\R^p$, the above procedure can be easily generalized to get 
\begin{equation}
\label{eqn:fkpconstbeta}
\begin{aligned}
\partial_t p(\bw,t) & =  \sum_{i=1}^p\frac{\partial}{\partial w_i}\left[\nabla L(\bw)p(\bw,t)\right] +\sum_{i=1}^p \frac{\partial^2}{\partial w_i^2}\left[\frac{\gamma(t)\beta}{2M(t)}p(\bw,t)\right]\\
& = \nabla\cdot\left(\nabla L(\bw)p +\frac{\gamma(t)\beta}{2M(t)}\nabla p\right).
\end{aligned}
\end{equation}
Since $\bW(0)=\bw_0$, $p(\bw,0)=\delta(\bw_0)$. This completes the derivation of the Fokker-Planck equation for constant $\beta(\bw) =\beta$.
For deriving (\ref{eqn:probdensitypthetatw}), we can apply (\ref{eqn:fkpconstbeta}) and notice that 
\begin{equation*}
\nabla\left[\frac{\gamma(t)\beta(\bw)}{2M(t)}p\right] = \nabla\left[\frac{\gamma(t)\beta(\bw)}{2M(t)}\right]p + \frac{\gamma(t)\beta(\bw)}{2M(t)}\nabla p.
\end{equation*}
This completes the proof.


\section{Discussion on the main assumptions (A.1) -- (A.3).}
\label{sec:disonassump}

We verify (A.1) and (A.2) for the $L_2$ loss and the mean cross entropy loss. 
Denote by $\{(\bx_n,y_n),1\leq n\leq N\}$ the set of training data. 
Without loss of generality, consider $\text{Var}[y_n|\bx_n] = 1$.

First, we consider the $L_2$ loss:
$L(\bw) = (\bw-\bw^0)^\top\E[\bx_n\bx_n^\top](\bw-\bw^0)+1$. By assumption that $\bssigma^2(\bw)$ is positive definite, we have
\begin{equation}
\label{eqn:a1lst}
\begin{aligned}
\lim_{\|\bw\|\to+\infty}L(\bw)  & \geq \lim_{\|\bw\|\to+\infty} \lambda_{\min}\{\E[\bx_n\bx_n^\top]\}\|\bw-\bw^0\|^2 +1\\
&  \geq \lim_{\|\bw\|\to+\infty} \lambda_{\min}\{\E[\bx_n\bx_n^\top]\}[\|\bw\|^2/2-\|\bw^0\|^2/2] +1 = +\infty,
\end{aligned}
\end{equation}
where $\lambda_{\min}\{\cdot\}$ denotes the minimal eigenvalue.  Note that
\begin{equation*}
\begin{aligned}
\int e^{-L(\bw)} d\bw  & = \int e^{-(\bw-\bw^0)^\top\E[\bx_n\bx_n^\top](\bw-\bw^0)-1}\\
&  \leq\int e^{- \lambda_{\min}\{\E[\bx_n\bx_n^\top]\}[\|\bw\|^2/2-\|\bw^0\|^2/2]-1} < +\infty,
\end{aligned}
\end{equation*}
This proves (A.1). To prove (A.2), we only need to note that $\|\nb L(\bw)\|^2/2 = 2 (\bw-\bw^0)^\top\{\E[\bx_n\bx_n^\top]\}^2(\bw-\bw^0)$ and $ \nb\cdot\nb L(\bw) = \text{Tr}\{\E[\bx_n\bx_n^\top]\}$, and similarly to (\ref{eqn:a1lst}) we can prove 
\begin{equation*}
\lim_{\|\bw\|\to +\infty}  \left\{\|\nb L(\bw)\|^2/2 - \nb\cdot\nb L(\bw)\right\} = + \infty,
\end{equation*} 
and 
\begin{equation*}
\lim_{\|\bw\|\to +\infty}  \left\{\nb\cdot\nb L(\bw)/\|\nb L(\bw)\|^2\right\} = 0.
\end{equation*} The assumption (A.3) can be verified straightforwardly as (A.2).

Second, we consider the mean cross entropy loss regularized  with the $l_2$ penalty for  logistic regression. Without loss of generality, we only consider the binary  classification: $L(\bw) = \E[-\by_n\log\widehat{\by}_n -(1-\by_n)\log(1-\widehat{\by}_n)]+\lambda\|\bw\|^2$ with $\widehat{\by}_n = 1/(1+e^{-\bw\cdot\bx_n})$. Note that
\begin{equation*}
\begin{aligned}
\lim_{\|\bw\|\to+\infty}L(\bw)  \geq \lambda\|\bw\|^2 = +\infty, \ \ 
\int e^{-L(\bw)} d\bw   \leq \int e^{-\lambda\|\bw\|^2}d\bw<+\infty.
\end{aligned}
\end{equation*}
This proves (A.1). To prove (A.2), note that $\nabla L(\bw) = \E[-\bx_n\by_n + \bx_n/(1+e^{-\bw\cdot\bx_n})]+2\lambda\bw$ and $-\nabla\cdot\nabla L(\bw) = \frac{e^{-\bw\cdot \bx_n}}{(1+e^{\bw\cdot\bx_n})^2}[2\P(y_n=1)-1]\text{Tr}(\bx_n\bx_n^\top) $. Since $\lambda\|\bw\|^2\to \infty$, we have that $\|\nb L(\bw)\|^2/2 - \nb\cdot\nb L(\bw)\to\infty $ and $\nb\cdot\nb L(\bw)/\|\nb L(\bw)\|^2\to0 $ as $\|\bw\|\to\infty$. 
Similarly, the assumption (A.3) can be verified as (A.2).
This completes the proof.

\section{Proof of Lemma \ref{lem:stationbeta}}
\label{sec:proflemstationbeta}
Let $\eta(t)= 2M(t)/[\gamma(t)\beta]$.
 By setting $\partial_tp=0$, it can be verified that $p_\infty(\bw) = \kappa e^{-\eta_\infty L(\bw)}$ satisfies 
 \begin{equation*}
 \nabla\cdot(\nabla L(\bw)p + \frac{1}{\eta(t)}\nabla p)=0.
 \end{equation*} 
 Since $\beta(\bw)\equiv\beta$ and the assumption (A.1) ensures that $e^{-\eta_\infty L(\bw)}$ is well-defined,
$p_\infty(\bw)$ is a stationary solution.


\section{Proof of Theorem \ref{thm:sigm4.1proof}}
\label{sec:proofofthm:sigm4.1proof}

Parallel to the notation of $p_\infty(\bw) =  \kappa e^{-\eta_\infty L(\bw)}$, we let 
\begin{equation*}
\hp(\bw,t)\equiv \k(t) e^{-\eta(t)L(\bw)},
\end{equation*} 
where $\eta(t) = \frac{2M(t)}{\gamma(t)\b(t)}$ and $\k(t)$  is a time-dependent normalization factor such that $\int\hp(\bw,t)d\bw=1$. 
Observe that (\ref{eqn:probdensitypthetatw}) can be written as
\begin{equation}
	\pt_tp = \frac{1}{\eta}\nb_\bw\cdot\(\hp \nb_\bw\(\frac{p}{\hp}\)\).
	\label{eqn: fp_iso}
\end{equation}
Let $\hp$ be $\hp(t,\bw)  = \pinf(\bw)\d(t,\bw)$, where $\d(t,\bw)\equiv \frac{\kappa(t)}{\kappa}e^{L(\bw)\({\eta_\infty} - {\eta(t)}\)}$.
Denote by $h$ the scaled distance  from $p$ to $\pinf$:
\begin{equation*}
	h\equiv\frac{p - p_\infty}{\sqrt{p_\infty}},
\end{equation*}
then h satisfies the following equation,
\begin{equation}
\label{eqn:pertsolu}
\begin{aligned}
	\pt_t h =& \frac{1}{\eta\sqrt{\pinf}}\nb_\bw\cdot\l[\hp \,\nb_\bw\(\frac{1}{\d} + \frac{h}{\sqrt{\pinf} \d} \)\r]\\
	=& \frac{1}{\eta\sqrt{\pinf}}\nb_\bw\cdot\l[\pinf \(\nb_\bw L\hd + \nb_\bw L\hd\(\frac{h}{\sqrt{\pinf}}\) +\nb_\bw\(\frac{h}{\sqrt{\pinf}}\) \)\r],
\end{aligned}
\end{equation}
where $\hd(t) =\eta(t) - \eta_\infty$. 
Multiplying $h$ to the both sides of (\ref{eqn:pertsolu}) and integrating it over $x$, after integration by parts, one has, 
\begin{align}
	\frac{1}{2}\pt_t\ll h \rl^2 =& \frac{\hd}{\eta} \underbrace{\int \frac{h}{\sqrt{\pinf}}\nb_\bw\cdot\(\pinf \nb_\bw L\r)d\bw}_{I} + \frac{\hd}{\eta} \underbrace{\int\frac{1}{2}\ll\frac{h}{\sqrt{\pinf}}\rl^2 \nb_\bw\cdot\(\pinf\nb_\bw L\)d\bw}_{II}\nonumber\\
	&- \frac{1}{\eta} \underbrace{\int \pinf\ll\nb_\bw\(\frac{h}{\sqrt{\pinf}}\)\rl^2d\bw}_{III}.\nonumber
\end{align}
Note that 
\begin{equation*}
	\nb_\bw\cdot\(\pinf\nb_\bw L\) = \pinf\(\nb_\bw\cdot\nb_\bw L  - \eta_\infty\ll \nb_\bw L\rl^2 \),
\end{equation*}
so by Assumption $A3$, one has 
\begin{equation*}
	\lv\nb_\bw\cdot\(\pinf\nb_\bw L\)\rv \leq \pinf^{2/3}\max\{1 , \eta_\infty\}M,
\end{equation*}
which implies 
\begin{equation*}
	I \leq  \frac{\max\{1 , \eta_\infty\}M}{2}\(\ll h \rl^2 
	+ \int \pinf^{1/3}d\bw \).
\end{equation*}

For term $II$, first Assumption $A3$ is equivalent to 
\begin{equation}
\lim_{\ll \bw \rl\to \infty} \frac{\nb_\bw\cdot\nb_\bw L}{2\eta_\infty\ll \nb_\bw L\rl^2 } = 0.
\label{eqn: limw}
\end{equation} 
Furthermore, Assumption $A2$ and (\ref{eqn: limw}) implies that $\lim_{\ll \bw \rl\to\infty} \ll \nb_\bw L\rl^2 \to +\infty$, so there exists a constant $R$, such that
\begin{equation*}
	 \nb_\bw\cdot\nb_\bw L  - 2\eta_\infty\ll \nb_\bw L\rl^2 \leq \eta_\infty, \qd \eta_\infty\ll \nb_\bw L\rl^2 \geq \eta_\infty,\qd  \text{for }\forall \ll \bw \rl > R. 
\end{equation*}
Therefore one has,
\begin{equation*}
	 \nb_\bw\cdot\nb_\bw L  - \eta_\infty\ll \nb_\bw L\rl^2 \leq 0, \qd \text{for }\forall \ll \bw \rl > R.
\end{equation*}
By the continuity of the loss function, for $\ll \bw \rl \leq R$, there exists a constant $C_2$, such that
\begin{equation*}
	 \lv\nb_\bw\cdot\nb_\bw L  - \eta_\infty\ll \nb_\bw L\rl^2\rv \leq C_2, \qd \text{for }\forall \ll \bw \rl <R.
\end{equation*}
Combining the above two inequality gives the bound for term $II$,
\begin{equation*}
	\lv II \rv \leq \frac{C_2}{2}  \ll h \rl^2 . 
\end{equation*}

Thus combine the estimates for the term $I$ and $II$, one has
\begin{equation}
	I+II \leq  C_1 \ll h \rl^2 + C_1, \label{eqn:part2est}
\end{equation}
where $C_1 = \frac{1}{2}\max\{1 , \eta_\infty\}\max\l\{ \int \pinf^{1/3}d\bw, 1+\frac{C_2}{2}\r\}M$.

For term $III$, under Assumption $A2$, one has  the following Poincar\'e inequality (see, e.g., \citet{pavliotis2014stochastic}) on $\pinf d\bw$, 
\begin{equation*}
	  \int \ll\nb_\bw\(\frac{h}{\sqrt{\pinf}}\)\rl^2  \pinf \,d\bw  \geq C_P\int  \(\frac{h}{\sqrt{\pinf}} - \int h\sqrt{\pinf} d\bw \)^2 \pinf \,d\bw.
\end{equation*}
In addition,  the fact that $ \int h\sqrt{\pinf}\,d\bw = 0$ gives
\begin{equation}
	 III  \geq C_P\ll h \rl^2.\label{eqn:part1est}
\end{equation}
The reason why $ \int h\sqrt{\pinf}\,d\bw = 0$ comes from the conservation of mass. That is, if one integrates (\ref{eqn: fp_iso}) over $\bw$ and uses integration by parts, 
\begin{equation*}
	  \pt_t\(\int  p(\bw,t)\, d\bw\)  = 0, 
\end{equation*}
which implies $\int h\sqrt{\pinf} \,d\bw = \int p\, d\bw - \int \pinf \,d\bw = 0$.  So combining (\ref{eqn:part2est}) and (\ref{eqn:part1est}) gives, 
\begin{equation}
	  \frac{1}{2}\pt_t\ll h \rl^2 + \frac{C_P}{\eta} \ll h \rl^2 \leq \frac{C_1\hd}{\eta} \(\ll h \rl^2 +1\)
	  \label{eqn:energyest}
\end{equation}
Since $\eta(t) \to \eta_\infty >0$ as $t\to\infty$, so there exists $T$ large enough, such that for $\forall t>T$, 
\begin{equation}
	  \hd = \lv \eta(t) - \eta_\infty \rv \leq\min\l\{\frac{\eta_\infty}{3}, \frac{C_P}{3C_1}\r\}.
	  \label{eqn:condonT}
\end{equation}
Plugging $\hd \leq \frac{c}{2C_1}$ into (\ref{eqn:energyest}), one has
\begin{equation}
	  \frac{1}{2}\pt_t\ll h \rl^2 + \frac{2C_P}{3\eta} \ll h \rl^2 \leq \frac{C_P}{3\eta}, \qd \text{for }\forall t>T .
	  \label{mid_1}
\end{equation}
Futhermore, (\ref{eqn:condonT}) also implies $2\eta_\infty/3 \leq \eta(t) \leq 4\eta_\infty/3$, which indicates that 
\begin{equation*}
	 \frac{2C_P}{3\eta}\geq\frac{C_P}{2\eta_\infty}, \qd  \frac{C_P}{3\eta} \leq \frac{C_P}{2\eta_\infty}.
\end{equation*}
Therefore, (\ref{mid_1}) becomes
\begin{equation*}
	  \frac{1}{2}\pt_t\ll h \rl^2 + \frac{C_P}{2\eta_\infty} \ll h \rl^2 \leq \frac{C_P}{2\eta_\infty}, \qd \text{for }\forall t>T .
\end{equation*}
Integrate the above equation from $T$ to $t>T$, one has,
\begin{equation*}
	  \ll h(t) \rl^2 \leq \(\ll h(T) \rl^2 + \frac{C_P}{\eta_\infty}(t-T)\) - \frac{C_P}{\eta_\infty}\int_T^t  \ll h(s) \rl^2ds.
\end{equation*}
By Gronwall's Inequality, one ends up with, 
\begin{equation*}
	\ll h(t)\rl^2 \leq \(\frac{C_P}{\eta_\infty}(t-T) + \ll h(T) \rl^2\) e^{-\frac{C_P}{\eta_\infty}(t-T)}.
\end{equation*}

\begin{remark}
There are some work in the literature about the convergence of the Fokker-Planck equation solution. However, most of these results focus on the convex $L(\bw)$. See, e.g., \citet{arnold2001convex, pavliotis2014stochastic}.
These results are different from the case under our consideration.
\end{remark}

\section{Mathematical quantification of the constant $T$ in Theorem \ref{thm:sigm4.1proof}}
\label{sec:quantificT}

We note that $T$ should be large enough such that for all $t>T$, 
\begin{equation*}
	  \lv \eta(t) - \eta_\infty \rv \leq\min\l\{\frac{\eta_\infty}{3}, \frac{C_P}{3C_1}\r\}, \qd \text{where }C_1 = \frac{M}{2}\max\{1 , \eta_\infty\}\max\l\{ \int \pinf^{1/3}d\bw, 1+\frac{C_2}{2}\r\}.
\end{equation*}
Here $C_2>0$ is the bound for $\lv\nb_\bw\cdot\nb_\bw L  - \eta_\infty\ll \nb_\bw L\rl^2\rv$ in bounded domain $\{\ll \bw \rl <R\}$, such that
\begin{equation*}
	 \nb_\bw\cdot\nb_\bw L  - \eta_\infty\ll \nb_\bw L\rl^2 \leq \left\{
	 \begin{aligned}
	 &0, \qd \text{for }\forall \ll \bw \rl > R,\\
	 &C_2, \qd \text{for }\forall \ll \bw \rl <R.
	 \end{aligned}
	 \right.
\end{equation*}
This quantification of $T$ is based on the proof in Section \ref{sec:proofofthm:sigm4.1proof}.


\section{Proof of Theorem \ref{thm:probmainresult4111}}
\label{eqn:proofofthfinal}

Let 
\begin{equation*}
\begin{aligned}
P_\e(\check{\bw}) & = \P(\ll \bW_\gamma(\infty) - \check{\bw}\rl\leq \epsilon)
\end{aligned}
\end{equation*}
be the probability of $W(\infty)$ staying in the $\e$-neighborhood of global minimum $\check{\bw}$, and the probability density function of $W(\infty)$ is $\pinf$,  then
\begin{equation*}
\begin{aligned}
P_\e(\check{\bw}) & = \int_{\ll\bw - \check{\bw}\rl^2\leq\e^2}\kappa e^{-\eta_\infty L(\bw)}d\bw \\
& = \int_{\ll\bw - \check{\bw}\rl^2\leq\e^2} \kappa e^{-\eta_\infty[L(\check{\bw})+(\bw-\check{\bw})'\mathrm{\Delta}L(\check{\bw})(\bw-\check{\bw})+o\{(\bw - \check{\bw})^2\}]}d\bw
\end{aligned}
\end{equation*}
Since $\check{\bw}$ is a local minimum of $L(\bw)$, so $\mathrm{\Delta}L(\check{\bw})$ is positive definite,  then there exists an orthogonal matrix $O$ and diagonal matrix $\L$, such that $\mathrm{\Delta}L  = O' \L O$. For simplicity, we assume $\mathrm{\Delta}L = \L = \text{diag}(\lam_1, \cdots, \lam_d)$. 
\begin{equation*}
\begin{aligned}
\lim_{\e\to0}P_\e(\check{\bw})& = \lim_{\e\to0}\left[\kappa e^{-\eta_\infty L(\check{\bw})} \int_{\ll \bw \rl^2\leq \e^2} \prod_{j=1}^d e^{-\eta_\infty \lam_j w_j}    d\bw\right]      e^{\eta_\infty\epsilon^2}\\
&= \lim_{\e\to0}\left[\kappa e^{-\eta_\infty L(\check{\bw})}\prod_{j=1}^d\frac{1}{\sqrt{\eta_\infty\lam_j}}\int_{-\epsilon\sqrt{\eta_\infty \lam_j}}^{\epsilon\sqrt{\eta_\infty \lam_j}}e^{-w^2}dw\right]e^{\eta_\infty\epsilon^2}\\
&=\lim_{\e\to0}\l[\frac{\kappa e^{-\eta_\infty L(\check{\bw})}}{\eta_\infty^{d/2}}\prod_{j=1}^d\frac{1}{\sqrt{\lam_j}}\(\Phi\(\epsilon\sqrt{\eta_\infty \lam_j}\) - \Phi\(-\epsilon\sqrt{\eta_\infty \lam_j}\)\)\r]e^{\eta_\infty\epsilon^2},
\end{aligned}
\end{equation*}
where the first equality comes from change of variable $ \bw - \check{\bw} \to \bw$, and the second one comes from $\eta_\infty\lam_j\bw_j \to \bw_j$. Here $\phi(\cdot)$ in the last equality is the cumulative density function for standard normal distribution. Using the approximation of the cumulative density function in \citet{polya1945remarks}, one can simplify the above equation by
\begin{equation*}
\begin{aligned}
\lim_{\e\to0}P_\e(\check{\bw})  =&\lim_{\e\to0}\l[ \frac{\kappa e^{-2\eta_\infty L(\check{\bw})}}{\eta_\infty^{d/2}}\prod_{j=1}^d\sqrt{\frac{1-e^{-\e^2\eta_\infty\lam_j/\pi}}{\lam_j}}\r]e^{\eta_\infty\epsilon^2} \\
=& \lim_{\e\to0}\frac{\kappa e^{-2\eta_\infty L(\check{\bw})}}{\eta_\infty^{d/2}\text{det}(\mathrm{\Delta}L(\check{\bw}))}\l[e^{\eta_\infty\epsilon^2}\prod_{j=1}^d\sqrt{1-e^{-\e^2\eta_\infty\lam_j/\pi}}\r].
\end{aligned}
\end{equation*}
This completes the proof.

\


\section{Numerical illustrations of  Theorem \ref{thm:probmainresult4111}}
\label{sec:numerprob}
\begin{figure}
\centering
\includegraphics[width=1\textwidth]{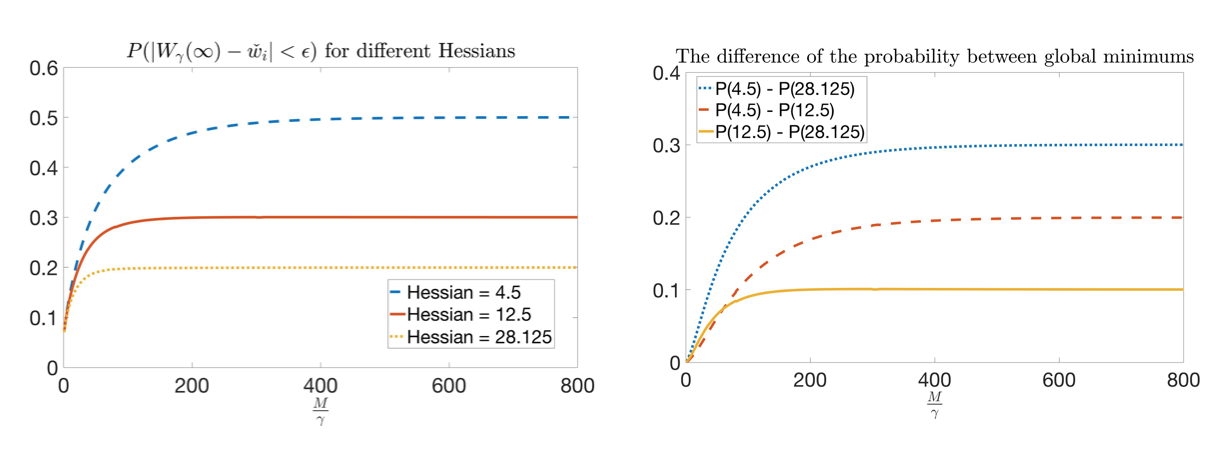}
\caption{Illustration of Example 1 with $\epsilon=0.1$. The left panel shows the probability of $W(\infty)$ staying in the $\epsilon$-neighborhood of different global minima. The right panel compares the differences of probabilities that $W(\infty)$  staying in the $\epsilon$-neighborhood of different global minima.}
\label{Hessian}
\end{figure}

\begin{figure}
\centering
\includegraphics[width=1\textwidth]{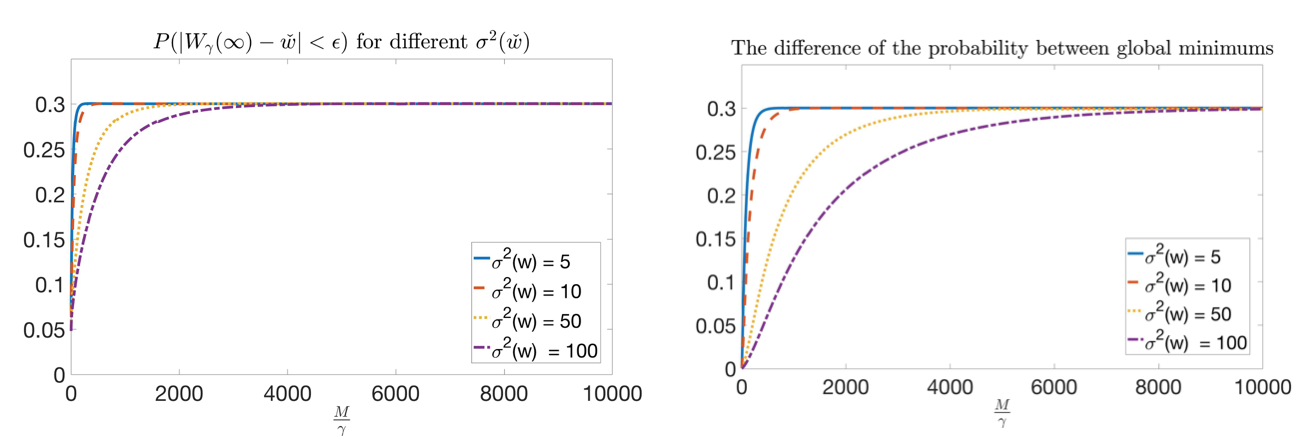}
\caption{Illustration of Example 1 with $\epsilon=0.1$. 
The left panel shows the probability of $W(\infty)$  staying in the $\epsilon$-neighborhood under different SGD variances $\s^2(\check{\bw})$. The right panel compares the differences of probabilities that $W(\infty)$ staying in the $\epsilon$-neighborhood of different $\s(\check{\bw})$.}
\label{sigma}
\end{figure}

To illustrate Theorem \ref{thm:probmainresult4111}, we explore three different examples showing how the probability changes with respect to $M/\gamma$,  $\mathrm{\Delta}L(\check{\bw})$, and the variance $\sigma^2(\check{\bw}) = \beta(\check{\bw})$:

\begin{itemize}
\item \textbf{Example 1}:  Consider the risk function $L(\bw)$ has three different global minima $\bw_i$, $i = 1,2,3$, with different Hessians $4.5$, $12.5$, and $28.125$, respectively. We are interested in the probability of the mini-batch SGD $\lim_{k\to\infty}\bw_k$ staying in the $\e$-neighborhood of global minima  with respect to the ratio $M/\gamma$, where $M$ is the batch size and $\gamma$ is the learning rate. The results are shown in Figure \ref{Hessian}.
\item \textbf{Example 2}:  Consider the variance of SGD has four different levels: $5, 10, 50, 100$. 
 We are interested in the probability of the mini-batch SGD $\lim_{k\to\infty}\bw_k$ staying in the $\e$-neighborhood of a same global minimum  of $L(\bw)$ with respect to the ratio $M/\gamma$.
 The results are shown in Figure \ref{sigma}.
\item \textbf{Example 3}: Consider two-dimensional cases. 
We are interested in the risk function $L(\bw)$ has two different global minima and furthermore, $L(\bw)$ has two different global minima. For two minima case, we consider $L(\bw)$ has two different Hessians $(2.42, 0.022)$ and $(2.22, 0.222)$. For three minima case, we consider $L(\bw)$ has three different Hessians $(15, 20)$, $(14.22, 42.66)$, and $(102.13, 25.53)$. 
\end{itemize}

\begin{figure}
\centering
\includegraphics[width=1\textwidth]{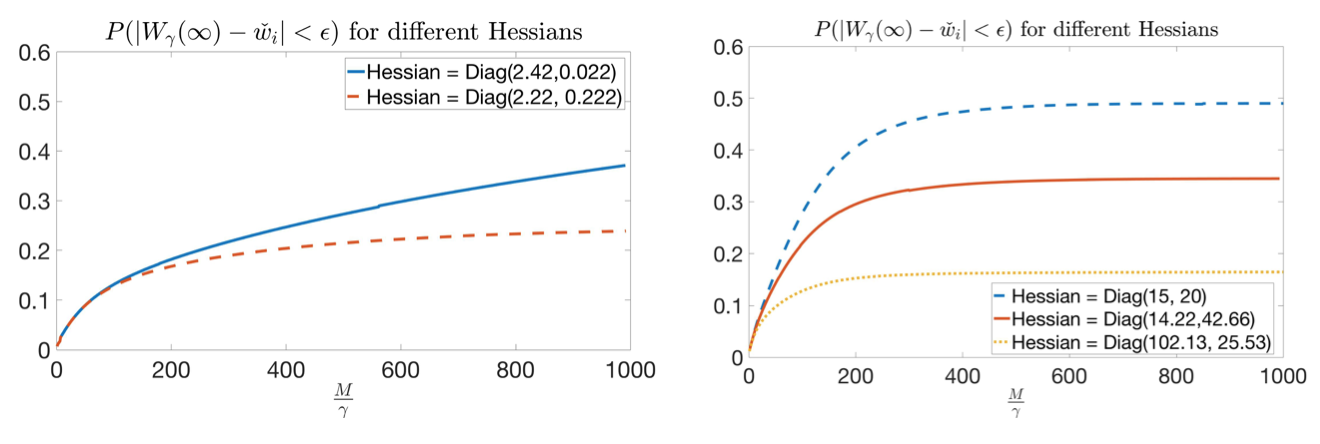}
\caption{Illustration of Example 3 with $\epsilon=0.1$. The left panel shows the probability of the limiting mini-batch SGD $\lim_{k\to\infty}\bw_k$ staying in the $\epsilon$-neighborhood of two different global minima. The right panel shows the probability of three different global minima. }
\label{twodim}
\end{figure}

\paragraph{Results.} 

The results of Example 1 are given in Figure \ref{Hessian}. We draw the following conclusions. 
\begin{itemize}
\item First,
if the batch size $M$ and learning rate $\gamma$ are the same, then $\bW(\infty)$ is more likely to stay near the flat minimum whose Hessian is smaller. 
\item Second, as the ratio $M/\gamma$ increases, the probability of $\bW(\infty)$ converging to a flatter minimum will increase faster than that of a sharper minimum. 
\item Third, if the ratio of Hessians (the Hessian at a sharp minima divides the Hessian at a flat minima) increases,  the difference of probabilities would increase as illustrated in the right panel of Figure \ref{Hessian}. Moreover,  if we increase the ratio $M/\gamma$, the difference  of probabilities becomes more distinct. 
\end{itemize}

 The results of Example 2 are given in Figure \ref{sigma}. We draw the following conclusion.
 \begin{itemize}
\item If the variance  $\s(\check{\bw})$ increases, the effect of the ratio $M/\gamma$ for the probability that converging to the global minimum will decrease. That implies as $\s(\check{\bw})$ increases, the probability of  SGD converging to a flat minimum will increase slower. 
 \end{itemize}

 The results of Example 3 are given in Figure \ref{twodim}. We draw the following conclusion.
 \begin{itemize}
 \item For a same ratio $M/\g$, if the product of the eigenvalues of the Hessian increases, then $\bW(\infty)$ will be more likely to stay near the minimum. For a same sharpness of the minimum, if one increase the batch size or decrease the learning rate, $\bW(\infty)$ will be more likely to stay near the minimum.
\item We conclude that the product of  eigenvalues of the Hessian matrix will affect the probability of $W(\infty)$  staying in the $\epsilon$-neighborhood of the minimum, which is different from the  sum of eigenvalues, the smallest eigenvalue, or the largest eigenvalue for multi-dimensional cases.
 \end{itemize}


\bibliographystyle{plainnat}

\begin{thebibliography}{18}
\providecommand{\natexlab}[1]{#1}
\providecommand{\url}[1]{\texttt{#1}}
\expandafter\ifx\csname urlstyle\endcsname\relax
  \providecommand{\doi}[1]{doi: #1}\else
  \providecommand{\doi}{doi: \begingroup \urlstyle{rm}\Url}\fi


 \bibitem[Anton. et~al.(2001)Anton, Markowich, Toscani, and
  \&~Unterreiter]{arnold2001convex}
Anton, A., Markowich, P., Toscani, G. \ \& Unterreiter, A. \ (2001)
 On convex Sobolev inequalities and the rate of convergence to
  equilibrium for Fokker-Planck type equations. {\it Communication in Partial Differential Equations}
  {\bf 26}(1--2):43-100.
  
\bibitem[Berglund(2013)]{berglund2013kramers}
Berglund, N. \ (2013)
Kramers' law: Validity, derivations and generalisations.
In {\it Markov Processes Relat. Fields}, {\bf 19}:459-490.

\bibitem[Bovier et~al.(2004)Bovier, Eckhoff, Gayrard and Klein]{bovier2004metastability}
Bovier, A., Eckhoff, M., Gayrard, V. \ \& Klein, M. \ (2004)
 Metastability in reversible diffusion processes I: Sharp asymptotics for capacities and exit times.
In {\it Journal of the European Mathematical Society}, {\bf 6}(4):399-424.

\bibitem[Bovier et~al.(2005)Bovier, Gayrard, Klein]{bovier2005metastability}
Bovier, A. Gayrard, V.  \ \& Klein, M. \ (2004)
Metastability in reversible diffusion processes II: Precise asymptotics for small eigenvalues.
In {\it Journal of the European Mathematical Society}, {\bf 7}(1):69-99.


\bibitem[Chaudhari et~al.(2017)Chaudhari, Oberman, Osher, Soatto, and  \&~Carlier]{Chaudhari2017deep}
Chaudhari, P., Oberman, A.,  Osher, S., Soatto, S. \ \& 
Carlier, G. \ (2017)
Deep Relaxation: partial differential equations for optimizing deep neural networks.
In {\it International Conference on Learning Representations}. 

\bibitem[Chaudhari et~al.(2017)Chaudhari and  \&~Soatto]{chaudhari2017stochastic}
Chaudhari, P. \ \& Soatto, S \ (2017)
Stochastic gradient descent performs variational inference, converges to limit cycles for deep networks
{\it arXiv preprint arXiv:1710.11029}. 

\bibitem[Choromanska et~al.(2015)Choromanska, Henaff, Mathieu, Arous, and
  LeCun]{Choromanska2015}
Choromanska, A., Henaff, M., Mathieu, M.,  Arous, G.B. \ \& 
LeCun, Y. \ (2015)
The loss surfaces of multilayer networks.
In {\it  Res Math Sci}, pp.\ 5--33. 




\bibitem[Dauphin et~al.(2014)Dauphin, Pascanu, Gulcehre, Cho, Ganguli, and
  Bengio]{dauphin2014}
Dauphin, Y.N., Pascanu, R., Gulcehre, C., Cho, K., Ganguli, S. \ \&
Bengio, Y. \ (2014)
Identifying and attacking the saddle point problem in
  high-dimensional non-convex optimization.
 In {\it Advances in Neural Information Processing Systems}, pp.\  2933--2941.

\bibitem[Dean et~al.(2012)Dean, Corrado, Monga, Chen, Devin, Mao, Senior,
  Tucker, Yang, Le, et~al.]{dean2012}
Dean, J., Corrado, G., Monga, R., Chen, K., Devin, M., Mao, M., 
Senior, A., Tucker, P., Yang, K., Le, Q.V. \ \& Ng, A.Y. \ (2012)
Large scale distributed deep networks.
 In {\it Advances in Neural Information Processing Systems}, pp.\
  1223--1231.

\bibitem[Dinh et~al.(2017)Dinh, Pascanu, Bengio, and Bengio]{Dinh}
Dinh, L., Pascanu, R., Bengio, S. \ \& Bengio, Y. \ (2017)
Sharp minima can generalize for deep nets.
In {\it International Conference on Machine Learning}.

\bibitem[Dziugaite and Roy(2017)Dziugaite and Roy]{dziugaite2017computing}
Dziugaite, G.K. \ \& Roy, D.M. \ (2017)
Computing nonvacuous generalization bounds for deep (stochastic) neural networks with many more parameters than training data.
{\it arXiv preprint arXiv:1703.11008}.


\bibitem[Ge et~al.(2015)Ge, Huang, Jin, and Yuan]{Ge2015}
Ge, R.,  Huang, R., Jin, C. \ \& Yuan, Y. \ (2015)
Escaping from saddle points--online stochastic gradient for tensor
  decomposition.
In {\it Conference on Learning Theory}, pp.\ 797--842.






\bibitem[Hoffer et~al.(2017)Hoffer, Hubara, and Soudry]{hoffer}
Hoffer, E., Hubara, I. \ \& Soudry, D. \ (2017)
Train longer, generalize better: closing the generalization gap in
  large batch training of neural networks.
In {\it Advances in Neural Information Processing Systems}, pp.\  1729--1739.


\bibitem[Jastrzebski et~al.(2017)Jastrzebski, Kenton, Arpit, Ballas, Fischer, Bengio, and Storkey]{Jastrzebski2017}
Jastrzebski, S., Kenton, Z., Arpit, D., Ballas, N., Fischer, A., Bengio, Y. \ \& Storkey, A. \ (2017)
Three factors influencing minima in SGD.
In {\it arXiv preprint arXiv:1711.04623}.


\bibitem[Keskar et~al.(2017)Keskar, Mudigere, Nocedal, Smelyanskiy, and
  Tang]{keskar}
Keskar, N.S., Mudigere, D., Nocedal, J., Smelyanskiy, M. \ \&
Tang, P.T.P. \ (2017)
On large-batch training for deep learning: Generalization gap and
  sharp minima.
In {\it International Conference on Learning Representations}.

\bibitem[Kolpas. et~al.(2007)Kolpas, Moehlis,  and
  \&~Kevrekidis]{kolpas2007coarse}
Kolpas, A., Moehlis, J. \ \& Kevrekidis, I.G. \ (2007)
 Coarse-grained analysis of stochasticity-induced switching between collective motion states. {\it Proceedings of the National Academy of Sciences}
  {\bf 104}(14):5931-5935.


\bibitem[LeCun et~al.(1998)LeCun, Bottou, Orr, and M{\"u}ller]{LeCun}
LeCun, Y., Bottou, L., Orr, G.B. \ \&  M{\"u}ller, K.R. \ (1998)
Efficient backprop.
In {\it Neural networks: Tricks of the trade}, pp.\ 9--50.
  Springer, Berlin, Heidelberg.

\bibitem[Li et~al.(2017)]{li2017}
Li, Q., Tai, C. \ \& E, W.  \ (2017)
Stochastic modified equations and adaptive stochastic gradient algorithms.
 {\it arXiv preprint arXiv:1511.06251}.
 
 \bibitem[Mandt et~al.(2017)Mandt, Hoffman, and
  Blei]{mandt2017}
Mandt, S., Hoffman, M.D. \ \&
Blei, D.M. \ (2017)
Stochastic gradient descent as approximate bayesian inference
In {\it Journal of Machine Learning Research}  {\bf 18}:1-35.
 
 
 \bibitem[Pavliotis(2014)]{pavliotis2014stochastic}
Pavliotis, G.A. \ (2014)
Stochastic processes and applications: diffusion processes, the Fokker-Planck and Langevin equations.
  Springer.
  
  \bibitem[Pavliotis(2018)]{pavliotis2014stochastic}
Pavliotis, G.A. \ (2014)
{\it Stochastic processes and applications: Diffusion processes, the Fokker-Planck and Langevin equations.}
Springer.
  
   \bibitem[Poggio(2017)]{poggio2017theory}
Poggio, T.,  Kawaguchi, K., Liao, Q., Miranda, B., Rosasco, L., Boix, X., Hidary, J.  \ \& Mhaskar, H. \ (2017)
Theory of Deep Learning III: explaining the non-overfitting puzzle.
{\it arXiv preprint arXiv:1801.00173}.
 
  
 
\bibitem[P{\'o}lya(1945)]{polya1945remarks}
P{\'o}lya, G. \ (1945)
Remarks on computing the probability integral in one and two dimensions.
In {\it Proceedings of the 1st Berkeley Symposium on Mathematical Statistics and Probability}

  
 \bibitem[Raginsky(2017)]{raginsky2017non}
Raginsky, M., Rakhlin, A. \ \& Telgarsky, M. \ (2017)
Non-convex learning via Stochastic Gradient Langevin Dynamics: a nonasymptotic analysis.
{\it arXiv preprint arXiv:1702.03849}.
 
 






  

 
\bibitem[Wu  and Zhu(2017)Wu and Zhu]{wu2017}
Wu, L..  \ \& Zhu, Z. \ (2017)
Towards Understanding Generalization of Deep Learning: Perspective of Loss Landscapes.
 {\it arXiv preprint arXiv:1706.10239}.
 
  
\bibitem[Zhang et~al.(2017)Zhang, Bengio, Hardt, Recht, and Vinyals]{zhang2017}
Zhang, C., Bengio, S., Hardt, M., Recht, B. \ \& Vinyals, O. \ (2017)
 Understanding deep learning requires rethinking generalization.
In {\it International Conference on Learning Representations}.


\end{thebibliography}

\end{document}